%% file: main.tex
\pdfoutput=1

\documentclass[11pt]{article}

\usepackage[final]{acl}

\usepackage{times}
\usepackage{latexsym}

\usepackage[T1]{fontenc}

\usepackage[utf8]{inputenc}

\usepackage{microtype}

\usepackage{inconsolata}

\usepackage{graphicx}

\usepackage{booktabs}
\usepackage{hyperref}
\usepackage{rotating}

\title{Natural Language Processing for Human Resources: A Survey}

\author{Naoki Otani \quad Nikita Bhutani \quad Estevam Hruschka \\
         Megagon Labs \\ \texttt{\{naoki,nikita,estevam\}@megagon.ai}}

\begin{document}
\maketitle
\begin{abstract}

Advances in Natural Language Processing (NLP) have the potential to transform HR processes, from recruitment to employee management. While recent breakthroughs in NLP have generated significant interest in its industrial applications, a comprehensive overview of how NLP can be applied across HR activities is still lacking. This paper discovers opportunities for researchers and practitioners to harness NLP's transformative potential in this domain. We analyze key fundamental tasks such as information extraction and text classification, and their roles in downstream applications like recommendation and language generation, while also discussing ethical concerns. Additionally, we identify gaps in current research and encourage future work to explore holistic approaches for achieving broader objectives in this field.

\end{abstract}

\section{Introduction}

Human Resources (HR) is a vital component of any organization, responsible for managing its most valuable resource---people. Over the years, computational tools have transformed HR processes like hiring, training, and administration, reshaping the labor market and workplace. At the same time, concerns about the accuracy and fairness of automated systems have also garnered significant attention, paving the way for ongoing and future research. Advancements in Natural Language Processing (NLP), especially with large language models (LLMs), have spurred interest in applying language technologies to a broad range of real-world problems, and the HR domain is no exception. However, this domain remains relatively underrepresented in the NLP research community.\footnote{Despite the development of many innovative applications in the industry~\cite{barth-2024-unveiling}, major conferences such as ACL, NAACL, EMNLP, EACL, and COLING featured only three papers with ``job'' or ``human resources'' in their titles in 2024.}

As breakthroughs in LLMs continue to advance various aspects of NLP, key challenges in the HR domain, such as the complexity of processing heterogeneous data, and the scarcity of publicly available data resources, may be alleviated in the coming years. Therefore, the HR domain holds substantial potential for growth and also presents unique challenges that can drive NLP research forward. To facilitate this transformation, it is essential to develop a comprehensive overview of key HR activities from an NLP perspective and examine how upstream tasks, such as skill extraction, contribute to downstream applications like job matching.

\begin{figure}[t]
    \centering
    \includegraphics[width=1\linewidth]{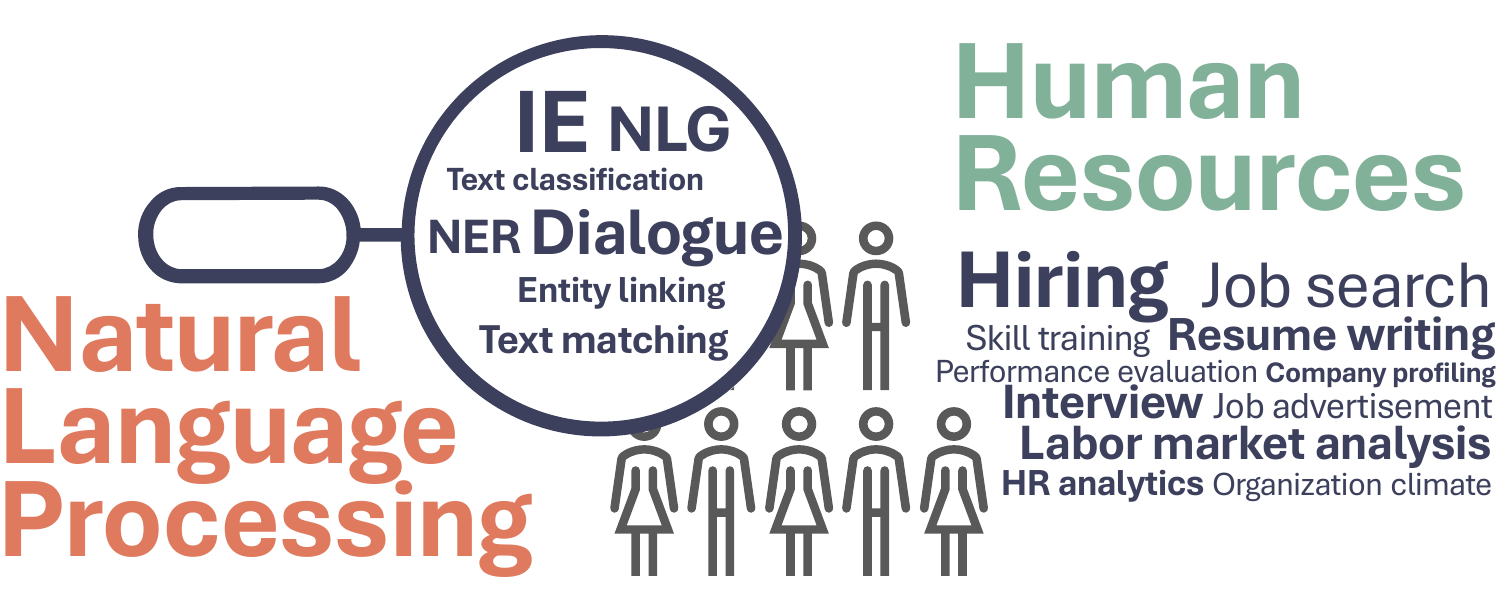}
    \caption{\textbf{Concept of this survey paper.} We review and categorize HR-related problems through the lens of core NLP research areas.}
    \label{fig:concept}
\end{figure}

\begin{figure*}[t]
    \centering
    \includegraphics[width=1\linewidth]{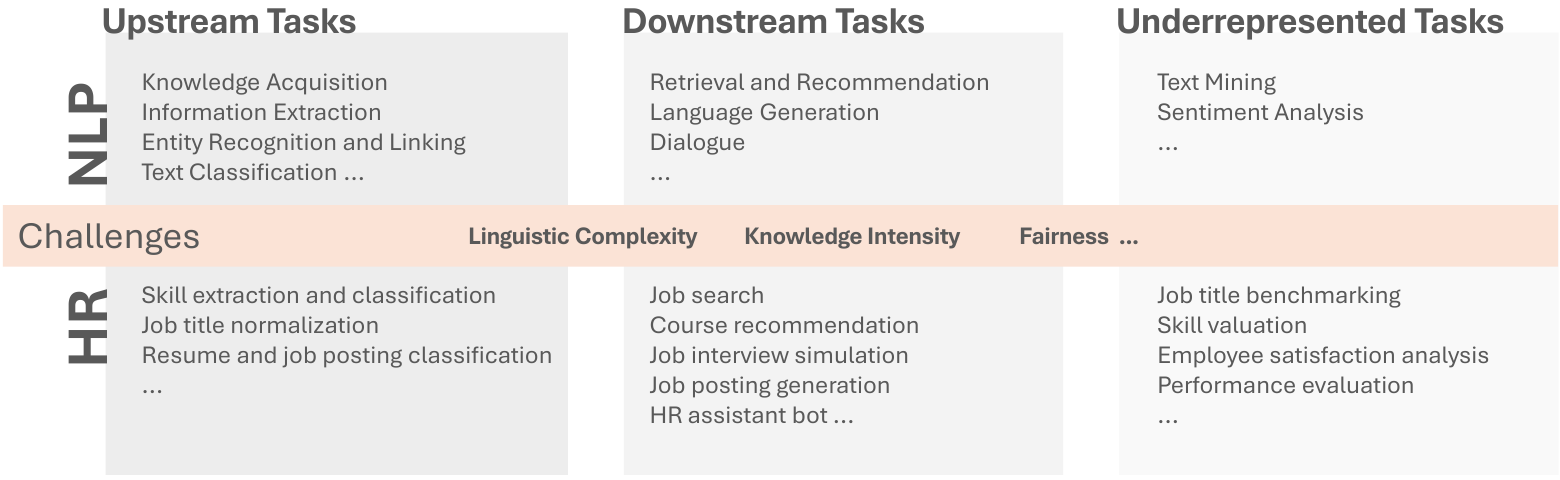}
    \caption{Landscape of NLP applications within the HR domain.}
    \label{fig:overview}
\end{figure*}

In this paper, we analyze HR activities through the lens of NLP research, categorizing them into key areas and examining how NLP techniques have been applied, along with remaining challenges~(Figure~\ref{fig:concept}).\footnote{NLP research is relevant to various HR activities. However, most existing studies focus primarily on talent acquisition, which is why this topic receives greater emphasis in the paper.} We explore fundamental tasks like information extraction and text classification~(\S\ref{sec:upstream-tasks}), and their role in supporting core applications such as recommendation, language generation, and interaction~(\S\ref{sec:downstream-tasks}). Finally, we highlight underrepresented areas in NLP to guide future research~(\S\ref{sec:underrepresented-tasks}). By organizing the discussion around NLP research topics, our goal is to provide insights for two audiences: (1) NLP researchers seeking impactful problems in the HR domain, and (2) those exploring how NLP can address HR challenges.

Previous surveys on this topic have typically focused on specific HR tasks and applications, such as information extraction from job postings~\cite{khaouja-etal-2021-survey,senger-etal-2024-deep}, market analysis~\cite{rahhal-etal-2024-data}, job recommendation~\cite{balog-etal-2012-expertise,de-ruijt-bhulai-2021-job,freire-etal-2021-e,mashayekhi-etal-2024-challenge}, conversational agents~\cite{laumer-morana-2022-hr}, and fairness~\cite{hunkenschroer-and-luetge-2022-ethics,kumar-etal-2023-fairness,fabris-etal-2024-fairness}. While general literature reviews in this field provide a broad overview of relevant computational research~\cite{budhwar-etal-2022-artificial,sharma-2021-literature,qin-etal-2023-comprehensive,khan-2024-application}, they do not specifically explore insights into language technologies. In contrast, we focus on core NLP methodologies, such as information extraction, text classification, retrieval, and language generation, and discuss their evolving role in various HR applications.\footnote{While a position paper by \citet{leidner-stevenson-2024-challenges} also explores NLP applications in this field, it does not provide a comprehensive literature review.}

This paper provides a structured NLP-centric perspective that systematically maps NLP tasks to HR challenges, making it easier for readers with an NLP background to identify relevant research opportunities and for HR practitioners to connect with relevant methods.\footnote{We describe our literature survey methodology in Appendix \ref{app:paper-collection}.} We highlight how specialized tasks contribute to broader goals, such as job title understanding for skill extraction and skill extraction for job matching, and encourage future work to explore holistic approaches for achieving these objectives. To further advance this field, we recommend the development and use of real or real(istic) datasets to enhance the relevance and impact of research outcomes.

\input{sections/scenarios}

\begin{figure*}[t]
    \includegraphics[width=1\linewidth]{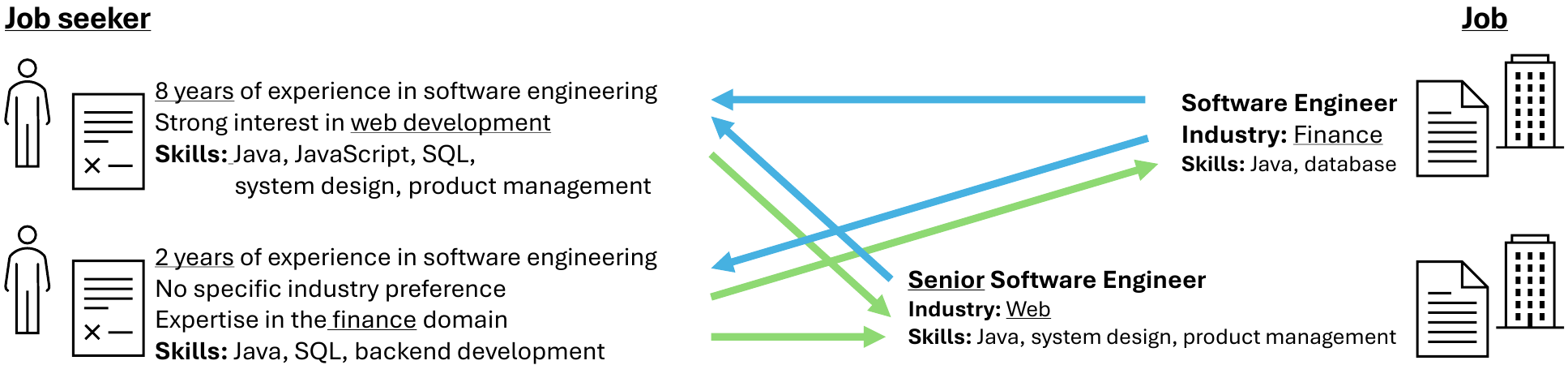}
    \caption{The problem of \textbf{job recommendation} (\S\ref{sec:information-retrieval}) is a two-sided process relying on multiple facets of information, such as expertise and requirements. Even if a job seeker prefers a particular job, the candidate may not necessarily be the best fit for the position.}
    \label{fig:job-recommendation-example}
\end{figure*}

\input{sections/upstream-tasks}

\input{sections/downstream-tasks}

\input{sections/underrepresented-tasks}

\input{sections/conclusion}

\section*{Limitations}

Due to space constraints, this paper aims to provide a focused literature review to offer readers a concise yet effective overview of HR applications. For those interested in a broader collection of NLP research in HR, we provide a list of papers and language resources on GitHub,\footnotemark[10] which we plan to update regularly. While there are numerous other NLP challenges in HR, such as linguistic and societal analysis (e.g., demographic, language, and cultural differences), we did not extensively cover these topics due to space limitations. As a result, the majority of papers discussed focus on widely spoken languages like English and Chinese. Lastly, while many companies are adopting modern NLP solutions in HR tasks, we have only reviewed techniques published in academic conferences.

\section*{Acknowledgments}

We thank the anonymous reviewers, our colleagues at Megagon Labs, and the participants of the NLP4HR 2024 workshop for their valuable feedback and discussions.

\bibliography{ref}

\appendix

\input{sections/appendix}

\end{document}

%% file: sections/scenarios.tex
\section{What is HR Concerned with?}\label{sec:scenarios}

This section briefly describes HR activities and their links to NLP. Broadly, these activities can be categorized into pre-hiring and post-hiring tasks.

\paragraph{Pre-hiring:} The pre-hiring process for recruiters includes drafting job postings, selecting candidates, conducting interviews, and extending offers. For job seekers, it involves exploring market trends, pursuing necessary training, preparing resumes, applying for jobs, preparing for interviews, and negotiating offers. These tasks rely heavily on nuanced domain-specific knowledge and are closely related to language generation (e.g., writing job postings and resumes, text-based communication) and specialized dialogue (e.g., interviews).

\paragraph{Post-hiring:} Key HR functions include setting role requirements aligned with organizational goals, evaluating performance, optimizing team dynamics, and maintaining positive work environments. These tasks are complex, demanding occupation-specific insights and integration of diverse data sources like employee records, organizational network, and textual communications.

The application of NLP techniques for these activities faces several challenges: (1)
\textbf{Diverse entities and language expressions} in HR data, such as the concise, bullet-pointed style of resumes or performance feedback, which vary across industries (e.g., software development vs. culinary arts). (2) The need for deep understanding of  
\textbf{domain-specific knowledge}, which is often not readily available in raw text corpora %
. (3)  \textbf{Biases in data-driven systems}, reflecting stereotypes, proxies for sensitive attributes, and external barriers~\cite{calanca-etal-2019-responsible,glazko-etal-2024-identifying}.

The following sections review existing research on HR activities, organized by NLP topics, with a focus on upstream tasks (\S\ref{sec:upstream-tasks}) and downstream tasks (\S\ref{sec:downstream-tasks}). We then discuss underrepresented HR activities (\S\ref{sec:underrepresented-tasks}) that could benefit from recent NLP advancements~(Figure~\ref{fig:overview}).

%% file: sections/upstream-tasks.tex
\section{Upstream Tasks}\label{sec:upstream-tasks}

Upstream HR tasks aim to enrich raw text corpora through information extraction and classification to facilitate knowledge-intensive downstream tasks.

\subsection{Taxonomy Creation}\label{sec:knowledge-acquisition}

Significant efforts have been made to acquire domain-specific knowledge and develop HR-related taxonomies to organize information on occupations, industries, skills, education, and certifications. This has led to the creation of large-scale resources such as the 
European Skills, Competences, Qualifications and Occupations (ESCO; \citealp{le-vrang-etal-2014-esco}) and others~\cite{lau-and-sure-2002-introducing,ilo-2012-international,bastian-etal-2014-linkedin,onet}. Expert-driven taxonomy creation can yield high-quality resources, but maintaining them is challenging. To reduce the costs, some studies have used Wikipedia~\cite{kivimaki-etal-2013-graph,zhao-etal-2015-skill} and the consolidation of web resources~\cite{gugnani-and-misra-2020-implict}. However, taxonomy creation remains highly complex due to cultural and regional variations~\cite{tu-cannon-2022-beyond}.

\subsection{Information Extraction}\label{sec:information-extraction}

The extraction of HR-related information, particularly job-related skills, has been extensively studied in the research community~\cite{khaouja-etal-2021-survey,senger-etal-2024-deep}. Skills include a range of competencies, such as technical expertise, knowledge, and the ability to learn and apply new concepts~\footnote{Some literature differentiates skills from knowledge, competencies, and qualifications, but for simplicity, we consider skills to encompass all types of proficiency.}. Other studies have also focused on extracting information like work experience and education~\cite{de-sitter-and-daelemans-2003-information,finn-and-kushmerick-2004-multi,green-etal-2022-development}.

This challenge is often framed as a sequential labeling problem with models trained on in-domain corpora~\cite{sayfullina-etal-2018-learning,green-etal-2022-development,zhang-etal-2022-skillspan}. Recent studies have explored multi-task and transfer learning~\cite{fang-etal-2023-recruitpro,zhang-etal-2023-escoxlm-r,zhang-etal-2024-nnose} to address the diversity and long-tail nature of job-related information. For extraction from resumes, the use of layout information has proven useful. Early work by \citet{yu-etal-2005-resume} introduced a two-pass model that segments and labels resume sections before identifying specific details. A similar approach is adopted by \citet{yao-etal-2023-resuformer} for extracting information from resumes in PDF format.

\subsection{Classification and Entity Linking}\label{sec:classification-and-entity-linking}

\textbf{Classification of job-related documents} plays a crucial role, especially in hiring, by organizing the large volumes of content generated by job seekers and recruiters. Previous research has focused on classification tasks such as categorizing resumes by job type~\cite{inoubli-brun-2022-dgl4c} and sorting job postings into occupation categories~\cite{lake-2022-flexible}. Text classification within documents---such as detecting section types~\cite{wang-etal-2022-machop} or analyzing work experience details~\cite{li-etal-2020-competence-level}---can also be useful for downstream applications like job recommendation. Automated text classification methods have already been widely used in society as part of Applicant Tracking Systems~(ATS), which has also drawn attention to their potential bias issues~(\S\ref{sec:ethics-bias-and-fairness}).

\textbf{Job title normalization} involves consolidating job titles expressed into a finite set of occupation categories. Prior work has addressed this by incorporating skill information~\cite{decorte-etal-2021-jobbert} and behavioral data into computational modeling~(\citealp{liu-etal-2020-ipod,ha-etal-2020-improving}; \textit{inter-alia}). For example, \citet{zhang-etal-2019-job2vec} integrated a job transition graph to model compositional meaning of job titles, while \citet{zhu-hudelot-2022-towards} enhanced this graph by further adding edges from component words. Recent studies have demonstrated the effectiveness of Transformer-based text encoders~\cite{yamashita-etal-2023-james,laosaengpha-etal-2024-learning}, yet this task remains challenging due to issues such as the length of documents, the presence of irrelevant information (e.g., location), and the reliance on domain knowledge.

A similar task is \textbf{skill classification}, which involves mapping texts to a pre-defined taxonomy like ESCO (\citealp{le-vrang-etal-2014-esco}). Some studies have employed methods based on similarity matching, while others have formulated the task as a multi-label classification problem~\cite{senger-etal-2024-deep}. A notable challenge in this task is handling diverse skill labels.\footnote{For example, ESCO v1.2.0 contains 13,939 skills.} \citet{zhang-etal-2024-entity} demonstrated that entity linking models trained on Wikipedia data can be effectively adapted to the HR domain. Other studies have explored implicit relationships between occupations and skills to improve skill identification. \citet{bhola-etal-2020-retrieving} 
used a bootstrapping technique leveraging skill co-occurrence, while \citet{goyal-etal-2023-jobxmlc} used a job-skill graph to capture implicit relationships between skills. To collect training data efficiently, \citet{decorte-etal-2022-design} proposed distant supervision, and recent studies have used LLMs to synthesize annotated texts~\cite{decorte-etal-2023-extreme,clavie-soulie-2023-large,magron-etal-2024-jobskape}.

\subsection{Summary}

Upstream HR tasks face challenges such as language complexity and diversity, varying types of data, and insufficient labeled data for training. While existing research has introduced innovative approaches to address these issues, some challenges remain underexplored. These include handling implicit information (e.g., inferring job requirements like a ``driver's license'' for truck drivers) and scaling extraction methods to accommodate emerging jobs and skills.

%% file: sections/downstream-tasks.tex
\section{Downstream Tasks}\label{sec:downstream-tasks}

Downstream HR applications broadly leverage NLP techniques across retrieval and recommendation, language generation, and dialogue systems. This section delves into these areas, followed by a discussion on the challenges of fairness and bias within these tasks.

\subsection{Retrieval and Recommendation}\label{sec:information-retrieval}

\textbf{Job recommendation} (or Person-Job fit) is typically framed as a text matching problem between job descriptions and resumes, addressed by various encoding methods such as word/document vectors~\cite{elsafty-etal-2018-document,zhu-etal-2018-person,mogenet-etal-2019-predicting} and Transformers~\cite{lavi-etal-2021-consultantbert,mesut-and-toine-2023-exploration}.\footnote{For more comprehensive review of this field, refer to specialized survey papers~\cite{balog-etal-2012-expertise,de-ruijt-bhulai-2021-job,freire-etal-2021-e,mashayekhi-etal-2024-challenge}.} The task is inherently two-sided, requiring consideration of the multifaceted preferences of both recruiters and job seekers~(Figure~\ref{fig:job-recommendation-example}). To address this problem, previous work has extracted and integrated fine-grained factors like skills~(\citealp{dave-etal-2018-combined,li-etal-2020-deep,yao-etal-2022-knowledge}; \textit{inter-alia}), experience levels~\cite{li-etal-2020-competence-level}, and more~\cite{ha-thuc-etal-2016-search,luo-etal-2019-resumegan,gutierrez-etal-2019-explaining,lai-etal-2024-knowledge} into matching models. Leveraging the linguistic capability of LLMs is an emerging research area, with studies exploring how LLMs can refine documents to alleviate the challenge of linguistic complexity~\cite{zheng-etal-2023-generative,du-etal-2024-enhancing} and integrating structured knowledge to improve accuracy and interpretability~\cite{wu-etal-2024-exploring}.

\textbf{Course recommendation} aims to help people bridge skill gaps by matching them with relevant courses from various data sources. Existing methods identify underlying factors using Transformer encoders~\cite{hao-etal-2021-recommending}, Bayesian variational networks~\cite{wang-etal-2021-personalized}, and generative adversarial networks~\cite{zheng-etal-2023-generative-learning}. Recently, LLM-based systems have emerged with modular components for upstream tasks like skill extraction, entity linking, and matching~\cite{frej-etal-2024-course}.

Retrieval and recommendation tasks in the HR domain are highly knowledge-intensive and often involve challenges associated with the heterogeneity of data sources such as documents and behavioral data. Although existing approaches have developed sophisticated methods to tackle these challenges, there remains substantial potential for integrating pre-trained language models to improve language comprehension~\cite{zhu-etal-2024-understanding}.

\subsection{Language Generation}\label{sec:generation}
\textbf{Generating job postings and resumes} is an impactful real-world application\footnote{For instance, on Indeed's platform, more than 750,000 employers have used an automated job posting generation system for approximately 2 million jobs as of July 2024~\cite{batty-2024-attract}.} that requires a nuanced understanding of job-specific skills across diverse work environments. Creating accurate job requirements, in particular, heavily relies on domain knowledge. \citet{liu-etal-2020-hiring} represented the relationships between skills, company size, and job titles using graphs, employing graph neural networks to generate job requirements. Similarly, \cite{qin-etal-2023-automatic} used a topic model to incorporate skill information into a job requirements generator. Other work has addressed job posting generation as a data-to-text task using a rule-based system~\cite{somers-etal-1997-multilingual} and a fine-tuned language model~\cite{lorincz-etal-2022-transfer}, with a focus on the fluency and adequacy of the generated texts.

\textbf{Generating interview questions} is also a knowledge-intensive task in the HR domain that can streamline the time-consuming candidate screening process. Automated systems have shown promise in generating questions based on the key requirements of a job position~\cite{shi-etal-2020-learning}. Beyond this, NLP technologies can assist in crafting personalized questions by leveraging contextual information~\cite{inoue-etal-2020-job,rao-s-b-etal-2020-automatic}, structured knowledge~\cite{su-etal-2019-follow}, or web search~\cite{qin-etal-2019-duerquiz,qin-etal-2023-automatic}. %

Language generation tasks in the HR domain present several characteristic challenges. For instance, these tasks often involve generating output based on lengthy inputs with mixed topics (e.g., job postings). Existing work has typically focused on simplified problem settings (e.g., inputs that have already been parsed into skill tags). Accurately and efficiently processing such complex inputs remains an open problem.

\subsection{Dialogue Systems}\label{sec:dialogue}

\textbf{Job interviews} present significant NLP research opportunities. Researchers have developed automated interviewing systems for communication skills training, which provide feedback through visualizations of user behavior~\cite{hoque-etal-2013-mach,rao-s-b-etal-2017-automatic} and adapt their interactions based on emotional states~\cite{anderson-etal-2013-tardis,hartholt-etal-2019-virtual,kawahara-2019-spoken}. Additionally, techniques for post-interview assessment have been proposed, combining various visual and audio features linearly~\cite{nguyen-etal-2014-hire,rao-s-b-etal-2017-automatic,naim-etal-2018-automated} or with advanced neural networks~\cite{hemamou-etal-2019-hirenet}. These studies have advanced the state of the art in processing multi-modal information, such as facial expressions, gestures, and speech. The rapid development of multi-modal LLMs could lead to new advancements in job interview systems. However, simply applying LLMs without domain-specific tuning can be ineffective, as a deep understanding of specialized knowledge is crucial for conducting meaningful conversations~\cite{li-etal-2023-ezinterviewer}.

Interactive systems can also be used for \textbf{managing HR-related inquiry}. A case study by \citet{malik-etal-2022-may} showed positive effects of chatbots on employee experiences in HR activities. Collecting interactive data in specialized domains is challenging, but \citet{xu-etal-2024-hr} demonstrated the effectiveness of LLMs to simulate interactions for post-hiring HR transactions.

\subsection{Ethics, Bias, and Fairness}
\label{sec:ethics-bias-and-fairness}
Fairness concerns in algorithmic hiring have been widely studied in various research fields~\cite{hunkenschroer-and-luetge-2022-ethics,kumar-etal-2023-fairness,fabris-etal-2024-fairness}, with bias mitigation techniques focusing on reducing disparities in algorithmic outcomes across sensitive groups. These techniques span multiple stages of system development and evaluation~\cite{quinonero-candela-etal-2023-disentangling}, including  biased keyword removal from input text~\cite{de-arteaga-etal-2019-bias}, balanced data sampling, internal representation adjustments~\cite{hauzenberger-etal-2023-modular,masoudian-etal-2024-effective}, and post-processing methods~\cite{geyik-etal-2019-fairness}.

The association between occupations and sensitive attributes has also been a significant focus in text representation and generation.  Studies have shown that word embeddings link gender pronouns with specific job titles, such as ``she'' with ``nurse'' and ``he'' with ``physician''~\cite{sun-etal-2019-mitigating}.
Similar gender biases are found in system-generated texts~\cite{sheng-etal-2019-woman,borchers-etal-2022-looking}. For example, \citet{wan-etal-2023-kelly} found that person names, which can serve as proxies for sensitive attributes, influence LLM-generated reference letters.
\citet{an-etal-2024-do} and \citet{nghiem-etal-2024-gotta} also report name-related biases in LLM-based hiring decisions, highlighting the need for careful consideration in these applications.

\citet{blodgett-etal-2020-language} conducted a literature review and argued the importance of carefully conceptualizing ``bias'' and grounding it in theories established outside of NLP. In the HR domain, fairness and bias have been extensively studied for decades~\cite{bertrand-mullainathan-2004-are}. This rich theoretical and empirical foundation could offer valuable insights to NLP research. A notable example is the bias evaluation framework by \citet{wang-etal-2024-jobfair}. This framework is informed by insights from labor economics, legal principles, and existing benchmarks, enabling a comprehensive and theoretically grounded evaluation of hiring decisions generated by LLMs.

\subsection{Summary}

Downstream HR tasks are highly knowledge-intensive and also necessitate ethical and safety considerations. Researchers have addressed these with advanced modeling techniques that leverage detailed information such as extracted skills. Looking ahead, the contextual understanding, and reasoning capabilities of modern LLMs present an opportunity to develop holistic approaches that integrate specialized modules to address overarching goals in downstream HR tasks.

%% file: sections/underrepresented-tasks.tex
\section{Underrepresented Tasks}
\label{sec:underrepresented-tasks}

Finally, we discuss HR activities that have been underrepresented in NLP research. Some of these tasks have received attention in broader research communities, but significant opportunities remain to leverage language resources for advancing computational methods.

\subsection{Data Analytics}

Analyzing the labor market~\cite{rahhal-etal-2024-data} can greatly benefit from data/text mining techniques. The insights gained can be valuable for policymakers, educators, and businesses.

\textbf{Job title benchmarking} involves matching job titles with equivalent expertise levels across different companies. Similarly, \textbf{job mobility analysis} focuses on identifying transferability between jobs while accounting for their specialties and work environments. These tasks are similar to the task of job title normalization~(\S\ref{sec:classification-and-entity-linking}) but require a deeper analysis of individual roles and organizations. For example, a company's industry and size often influence an employee's next career move. Therefore, previous work has developed methods to integrate diverse information linked to career trajectories with LSTMs~\cite{li-etal-2017-nemo}, multi-view learning~\cite{zhang-etal-2019-job2vec} and graph neural networks~\cite{zhang-etal-2021-attentive,zha-etal-2024-career}.

\textbf{The assessment of skill demand and value} is important not only for hiring but also for economic research~\cite{zhu-etal-2018-world,cao-etal-2021-occupational} and education~\cite{hao-etal-2021-recommending,patacsil-and-acosta-2021-analyzing}. While this area has not yet gained much attention within the NLP community, a variety of techniques have been explored in the broader research field. For instance, \citet{sun-etal-2021-market} introduced a neural model to break down job positions into required skills and assess their market value through salary prediction. \citet{cao-etal-2024-cross} proposed a graph encoder over a skill co-occurrence graph to capture demand-supply patterns in skill evolution. More recently, \citet{chen-etal-2024-job} developed a large-scale dataset for forecasting job-skill demand, which opens avenues for future research. Although these studies effectively utilize structured data, skills are often described by simple phrases that may not fully convey their true functions. For example, ``communication skills'' can differ significantly based on the context (e.g., schools vs. consulting firms). Future research could focus on extracting rich contextual information from textual data such as job postings to enhance the depth of analysis.

\subsection{Sentiment Analysis and Opinion Mining}
Sentiment about jobs and organizations can be collected through questionnaires or reviews from platforms like Glassdoor.\footnote{\url{https://www.glassdoor.com/}} This information has the potential to help organizations create work environments, boost productivity, and improve business outcomes~\cite{harter-etal-2002-business}.

\textbf{Employee satisfaction (job satisfaction)} analysis focuses on evaluating work environments and identifying areas for improvement based on employee feedback. \citet{moniz-and-de-jong-2014-sentiment} applied topic modeling to online employee reviews to uncover key themes related to the organization's future.  \citet{rink-etal-2024-aspect} approached this as an aspect-based sentiment analysis task, creating annotated datasets and fine-tuning transformer-based classifiers. While these studies highlight valuable use cases of sentiment analysis, addressing the diversity of job categories remains an open challenge.

\textbf{Company profiling} focuses on identifying the key characteristics of a company. Early work relied mainly on numerical data, but recent studies have successfully incorporated textual data for deeper insights~\cite{bajpai-etal-2018-aspect,lin-etal-2020-enhancing}. For example, \citet{lin-etal-2020-enhancing} proposed a model-based topic approach that integrates review texts with numerical data to perform both qualitative opinion analysis and quantitative salary benchmarking. %

\subsection{Summary}

This section highlighted several HR activities that offer significant opportunities to explore NLP techniques with heterogeneous data. In a similar vein, other core HR tasks, such as employee performance evaluation~\cite{ye-etal-2019-identifying,yu-etal-2023-large} and turnover analysis~\cite{teng-etal-2019-exploiting,gamba-etal-2024-exit}, also provide interesting challenges. Future efforts should focus on constructing publicly accessible datasets to drive advancements in this area. Applying LLMs to synthesize data or de-identify Personally Identifiable Information (PII) in real-world datasets could offer a promising solution to the problem of data scarcity. However, they should be used with caution, as issues such as amplifying biases~(\S\ref{sec:generation}) and exposing sensitive information from training data~\cite{carlini-etal-2021-extracting} remain.

%% file: sections/conclusion.tex
\section{Conclusion and Future Directions}\label{sec:conclusion}

In this paper, we have categorized critical research
challenges within the HR domain and identified significant opportunities for future exploration. To inspire future research in this domain and the broader
NLP community, we provide a list of papers and
public data resources on GitHub,\footnote{\url{https://github.com/megagonlabs/nlp4hr-survey}} which we plan
to update regularly.

\paragraph{Toward Broader Goals:} The HR domain encompasses a variety of specialized problems where NLP techniques have been successfully applied (e.g., skill extraction). These problems are often tied to broader goals, such as matching talent with appropriate job opportunities and optimizing employee productivity. For example, accurate skill extraction can significantly improve job recommender systems. To accurately extract this skill information, it is useful to perform semantic analysis of documents to identify relevant sections and understand job titles. Intermediate tasks like these can improve system performance in downstream applications and provide detailed information that can improve the fairness and transparency of final outcomes. The orchestration of specialized NLP tools to perform complex tasks is increasingly gaining the interest of the research community~(e.g., \citealp{schick-etal-2023-toolformer}). The HR domain would benefit from exploring holistic approaches, which could also provide research opportunities to push the boundaries of language technologies.

\paragraph{Knowledge Transfer:} Some successful research in the HR domain has introduced techniques and knowledge transferable to problems in other applications or domains. This trend is particularly evident in studies on job recommendation and bias mitigation, where the HR domain has established a strong position within the research community. We can also see similar knowledge transfer in some other specialized domains. For instance, the e-commerce domain has been one of the key drivers of multiple core NLP areas such as information extraction, sentiment analysis, and summarization. Promoting knowledge transfer to other domains will be key to conducting impactful NLP research in HR in the future.

\paragraph{Data Challenge:}  The availability of real or realistic datasets is a critical factor for advancing NLP research in the HR domain. Many types of HR documents involve privacy concerns that make them unsuitable for public release. However, approaches such as shared tasks with restricted data licenses, data donation,\footnote{FINDHR collected more than 1,100 CVs through donations (\url{https://findhr.eu/datadonation/}).} anonymization, and data synthesis could provide valuable resources to the research community. Moreover, working with real-world datasets would also help researchers identify system constraints and requirements in practical scenarios such as latency requirements, increasing the social impact of research artifacts.

\paragraph{The Application of LLMs:} The application of LLMs has gained popularity in the HR domain. While the collection and annotation of HR documents pose significant challenges, some studies have demonstrated the potential of LLMs to alleviate these issues. Furthermore, LLMs may introduce a new paradigm for many problems, offering substantial opportunities for researchers to generate innovative ideas that benefit both the HR domain and the broader research community.

%% file: sections/appendix.tex
\section{Paper Collection}
\label{app:paper-collection}
In this paper we aimed to offer a curated overview of key research challenges rather than a systematic and exhaustive literature review due to the page limit. Before curating papers, we employed the following approach to gather relevant papers. We begin by identifying recently published HR-related papers using keywords such as ``job,'' ``occupation,'' ``hiring,'' ``recruit,'' ``resume,'' ``HR,'' ``company,'' and ``skill'' from venues such as ACL conferences, KDD, CIKM, WWW, SIGIR, RecSys, AAAI, IJCAI, and relevant workshops. Additionally, we conduct keyword searches on Google Scholar and Semantic Scholar to collect non-computational papers. Subsequently, we employ snowball sampling from the citations of these papers to further gather relevant literature. We include peer-reviewed academic papers available as of December 2024 and exclude the others unless they are cited from multiple academic papers.